# MIVAR: Transition from Productions to Bipartite Graphs MIVAR Nets and Practical Realization of Automated Constructor of Algorithms Handling More than Three Million Production Rules

**Oleg O. Varlamov**

Moscow Automobile and Road State Technical University (MADI), Russia

OVarlamov@gmail.com

**Annotation.** The theoretical transition from the graphs of production systems to the bipartite graphs of the MIVAR nets is shown. Examples of the implementation of the MIVAR nets in the formalisms of matrixes and graphs are given. The linear computational complexity of algorithms for automated building of objects and rules of the MIVAR nets is theoretically proved. On the basis of the MIVAR nets the UDAV software complex is developed, handling more than 1.17 million objects and more than 3.5 million rules on ordinary computers. The results of experiments that confirm a linear computational complexity of the MIVAR method of information processing are given.

## Introduction

The problem of elaboration of intellectual systems is up-to-date and important. Elaboration of expert systems of new generation would allow automation of solving difficult problems. The MIVAR (Multidimensional Informational Variable Adaptive Reality) approach has provided an opportunity to offer new models and methods for data mining and management. The MIVAR technologies have been being elaborated in Russia for quite a long period of time. The first articles concerned some problems of graph theory and the elaboration of lineal matrix method of logic inference path finding on the adaptive net of rules [1-3]. Then, some work concerning the elaboration of MIVAR information space was done [4-5]. The most strictly formalized definition of the MIVARs can be found in papers [6-7]. Then the questions of the development [8-10] and the use of MIVARs for different simulators and instruction systems were discussed [11-22]. The fullest overview of theory and last achievements in MIVARs can be found in papers [4,6,10,15,18].

MIVAR nets allow the creation of new "General Problem Solver" the prototype of which is an UDAV (Universal Designer Algorithms Varlamov) programme complex. MIVAR nets remove restrictions existed before and, in fact, create expert systems of new generation which are capable to process millions of rules in acceptable time. MIVAR nets can be seen as a qualitative leap and transition to the new possibilities in information processing.

Let's consider the systems of artificial intelligence as active self-learning logically thinking systems. In the last century, technologies of expert systems creating were elaborated for particular narrow subject domains. This was due to the difficulties in formalizing of required subjects domains as to the fact that systems of logic conclusion could not process more than 20 rules (because an exhaustive search is a Nondeterministically polynomical). At the same time, "intellectual software packages" (ISP) were developed, allowing the automated solution of problems from different domains where the calculations and the construction of algorithms were required. Technologies of ISP are developing in MIVARs and service-oriented architectures.

The MIVAR approach unifies and develops achievements from different scientific domains: databases, computational problems, logic processing, and includes two main technologies:

1) *The MIVAR technology of information accumulation* – is a method of creating of global evolutional bases of data and rules (knowledge) with changeable structure based on the adaptive discrete MIVAR information space of unified representation of data and rules which bases on three main definitions: "Thing, Property, Relation".

2) *The MIVAR technology of information processing* - is a method of creation of the system of logic inference or "automatic construction of algorithms from modules, services and procedures" based on the active MIVAR net of rules with lineal computational complexity.

The MIVAR technology of information accumulation is designed for keeping any information with possible evolutional change of its structure and without any restrictions of its volume and the form of representation.

The MIVAR technology of information processing is designed for the processing of information, including logic inference, computational procedures and services.

In fact, MIVAR nets allow to develop production approach and to create an automatic learning logically thinking system. The MIVAR approach unifies and develops production systems, ontology, semantic nets, service-oriented architectures, multi-agent systems and other modern information technologies.

Currently, the software complex "UDAV" executes the search of logic inference and automatically constructs algorithms of problem solving controlled by the flow of entrance data. UDAV processes more than 1.17 million of variables and 3.5 million of rules. Software realization proves the lineal computational complexity of the search of logic inference on practice.

**Analyses of paradigms and models of information processing**

Traditionally, there exist the following paradigms and models of data processing: propositional calculus, predicate calculus, productions, semantic nets, Petri nets, ontology, and others. Production approach has important advantages. D.A. Pospelov wrote that the knowledge about the world can have double nature:

1) can contain the description of facts and events of the world that fix their presence or absence, and main relations and regularities containing these facts and events;

2) can contain procedural descriptions of how these facts and events should be manipulated and different goals of the system should be achieved [23].

Productions are generally described as "IF… THEN…". Some specialists in intellectual systems believe that the description of knowledge in the form of productions is universal: all the knowledge can be described in this form. Different rules, procedures, formulas or services can be represented in the system of productions. In fact, all causal statements can be reduced to productions. So, the use of production approach for the logic and computational processing of different data is reasonable and efficient.

To solve various practical problems in the sphere of computer sciences databases (storage) and logical computational data processing is required. Historically, the areas of logic inference and computational processing have been developing independently and successfully solved problems of various classes. To some extent, there existed a contradiction between these two approaches [2, 3, 4, 10-17, 23-32]. In addition, the problems of processing and storage of the data were divided.

Databases were used to store and to search required data, systems of logic inference – to process the information. In a result, these two domains hardly intersected, although there were quite a lot of propositions to unite all the functions of data storage and processing in the single system [4, 10, 23-32].

The MIVAR approach allows to unify in the single formalism both data storage (in the MIVAR information space – database) and processing (in the MIVAR logical-computing nets).

The MIVAR approach allows solving various scientific and practical problems. First, let's make an analysis of previously existed approaches and estimate their limits. Then, we'll move to the analyses of problems, achievements and perspectives in the domain of databases and the MIVAR information space of unified representation of data and rules.

### Possibilities and limitations of production approach

Production approach possesses some important advantages. D.A. Pospelov proposes 9 types of productions and the possibility of other types to exist is underlined [23]. There are examples of knowledge represented in the form of productions. Pospelov defines production system as an aggregate of productions, which can include productions of all listed types. There are some production constructions. The general form is

$$i, \Pi, P, A{=}{>}B, Q.$$

Here, **A=>B** is an ordinary production "if…, else…" called a production core. **P** characterizes external condition or applicability conditions of production determined by factors which are not included in **A**. Condition **P** allows to chose needed productions from all production with **A** in left part of the core. **Π** characterizes the sphere of the subject domain of the knowledge base, or pre-conditions of production applicability. These preconditions are not different from ***P*** but they form the formal system in the frames of which logic reasoning will be drawn. ***Q*** characterizes post-conditions of the production indicating the changes that need to be brought into the base of knowledge and the system of productions after implementing of this production [23, p. 134-135].

In general form, productions are rather rare. Horoshevsky marks out an intermediate "layer of rules" for which the research of different logic inference on rules systems is made [31, p.82-83].

Logic processing is understood as some conclusion lying in the base of human reasoning. Each process of drawing a conclusion depends on the examination of options. To increase the efficiency of process of logic inference is a central problem of all systems of the deductive conclusion [23, p. 79].

D.A. Pospelov has described two methods of logic inference: the *method of forward wave* and the *method of backward wave*. In the first method, the wave of the searching paths is propagating from all initial areas to the aim area. In the second, this wave is propagating from the aim area towards initial areas. The difference of two methods comes to different number of steps of the search. The combined method is frequently used when forward and backward waves meet, and the path from initial data towards the aim is being driven [23, p.85]. Picture 1 shows an example the graph of non-oriented "AND-OR" net [23, p. 83].

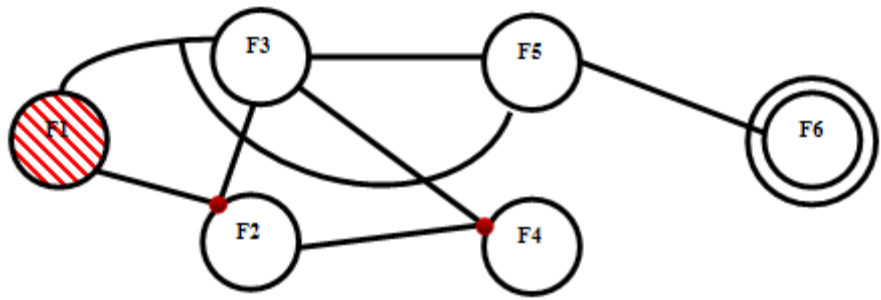

Picture 1. The graph of non-oriented "AND-OR" net.

These works didn't find wide practical using last century. That was caused by the fact of absence of the method of fast logic data processing. All methods used the exhaustive search, so the scale of real-time processing is about 20 rules while to solve problems processing of hundreds and more rules is required. The MIVAR approach and the creation of logic-computational nets have allowed to implement effectively the *methods of forward and backward waves* designed for productions by Pospelov [23, p.85].

**Artificial Intelligence systems and MIVAR**

Let's examine the approaches to the Artificial Intelligence and expert systems elaboration. Luger considered understood the intellect as a difficult area of knowledge that can't be described in the frames of one particular theory. The formal approach of Russel and Whitehead made it possible to automate it in real computational devices. Logical syntax and formal rules of inference elaborated by Russel and Whitehead are the base of automated proof of theorems systems and are the theoretic base of the AI [32].

Formalism, indeed, limits the possibilities of the description of real subject domains. A. Tarski's theory of reference played the dominant role in the formation of AI. According to this theory, correctly constructed formula of Frege and Russel-Whitehead refers to the objects of real world, and this conception lies in the basis of a majority of formal semantics theories. Luger notes that AI didn't manage to become a live scientific domain until the appearance of digital computational devices.

The architecture of digital computers forces the specific understanding of AI theory. The intellect appears to be a method of information processing. Next Luger formulates a wonderful idea, to which we are happy to join in: "We often forget that the tools that we create for their own purposes affect on the formation of our ideas about the world." This interaction is an important aspect of human knowledge: a tool (as scientific theories, ultimately, also the tools) is created to solve a specific problem. In this process the application and development the tool suggests other ways to use it, which lead to new questions and, ultimately, the development of new tools [32, p. 35].

If we talk about MIVARs and the formalisms Frege, Russell and Whitehead, Tarski, and many others, we must remember that science progresses in a spiral and the old formalisms require its continuation on a new coil. The above-mentioned predicate calculus and other early fundamental formalisms, if necessary can be implemented in the rules (relationships) of MIVAR netws. But we must remember that there are other formalisms for providing information for the AI, while the first order predicate calculus is fairly restrictive and not very powerful. We consider the creation of global learning-diagnostic systems that would solve complex logical and computational problems in real time. As can be seen from [23-32], in the modern theory of AI the predicate approach has already developed a large number of other approaches that differ from the predicate calculus. We emphasize that the production approach and its development in the MIVAR nets are another alternative to the predicate calculus to create AI.

### Representation of production nets in the form of bipartite graphs

Let's formalize productions and nets that can be formed on their basis. D.A. Pospelov [23] represents the net of rules in the form of a graph. Papers [1-22] offer an approach of correlation between nets of rules and graphs. Besides, it is shown that some problems of logic inference can be solved basing on the graph theory. Nets of rules and procedures are appropriate to represent in the form of bipartite graphs resulting in somewhat of Petri nets with the corresponding development in the MIVAR nets [1-22]. Let's remember the definition of bipartite graphs. "Graph $G=(V, E)$ is called bipartite if there is partition $V=\{V_1, V_2\}$ where none of two vertexes from $V_1$ or $V_2$ are adjacent" [25, p.223].

"Bipartite graph $G=(X, Y, E)$ is non-oriented graph all the vertexes of which can be divided into two classes $X$ and $Y$ so that the ends of each edge belong to different classes" [24, p.125]. That definition allows natural generalization. Non-oriented graph is called $k$-partite if its vertexes can be divided into $k$ classes so that the ends of each edge belong to different classes [24, p.125]. So, we can use bipartite, three-partite and $k$-partite graphs for different subject domains. It is a new approach towards data representation that has gained a name of the MIVAR approach. MIVAR approach is the generalization and development of production approach, Petri nets and other formalisms used in data processing.

### MIVAR nets

MIVAR nets can be represented as a bipartite graph that consists of the objects-variables and the rules-procedures. First of all, two lists forming two non-intersecting parts of the graph are being made: list of the objects and list of the rules. Objects are represented as ellipses (circles) at the Pic. 2. Each rule in the MIVAR net is a development of productions, hyper-rules with multi-activators or computational procedures. It is proved that all these formalisms are identical from the point of view of their following processing. In fact, these are the vertexes of the bipartite graph (which are represented as rectangles). Pic. 3 shows an example of the record of "objects" and "rules" of the MIVAR net in XML, and the algorithm of logic processing for the UDAV software complex is shown on the Pic. 4 [10-22].

### MIVAR method of logic and computational data processing

The following must be fulfilled in order to realize the method of logic and computational data processing. First, there must be a formalized description of subject domain. To make this formalization, main objects-variables and rules-procedures are being determined basing on the MIVAR approach. Then, the lists of "objects" and "rules" are being created. The formalized description of the method of logical and computational data processing is analogical to the bipartite graph of the MIVAR logic net, showed on Pic.2. Theoretic basis of the MIVAR method of logical and computational data processing were firstly published in the year 2002 [2-4]. The article [2] presents theoretic basis of creation of lineal matrix method of logic inference path on the adaptive net of rules.

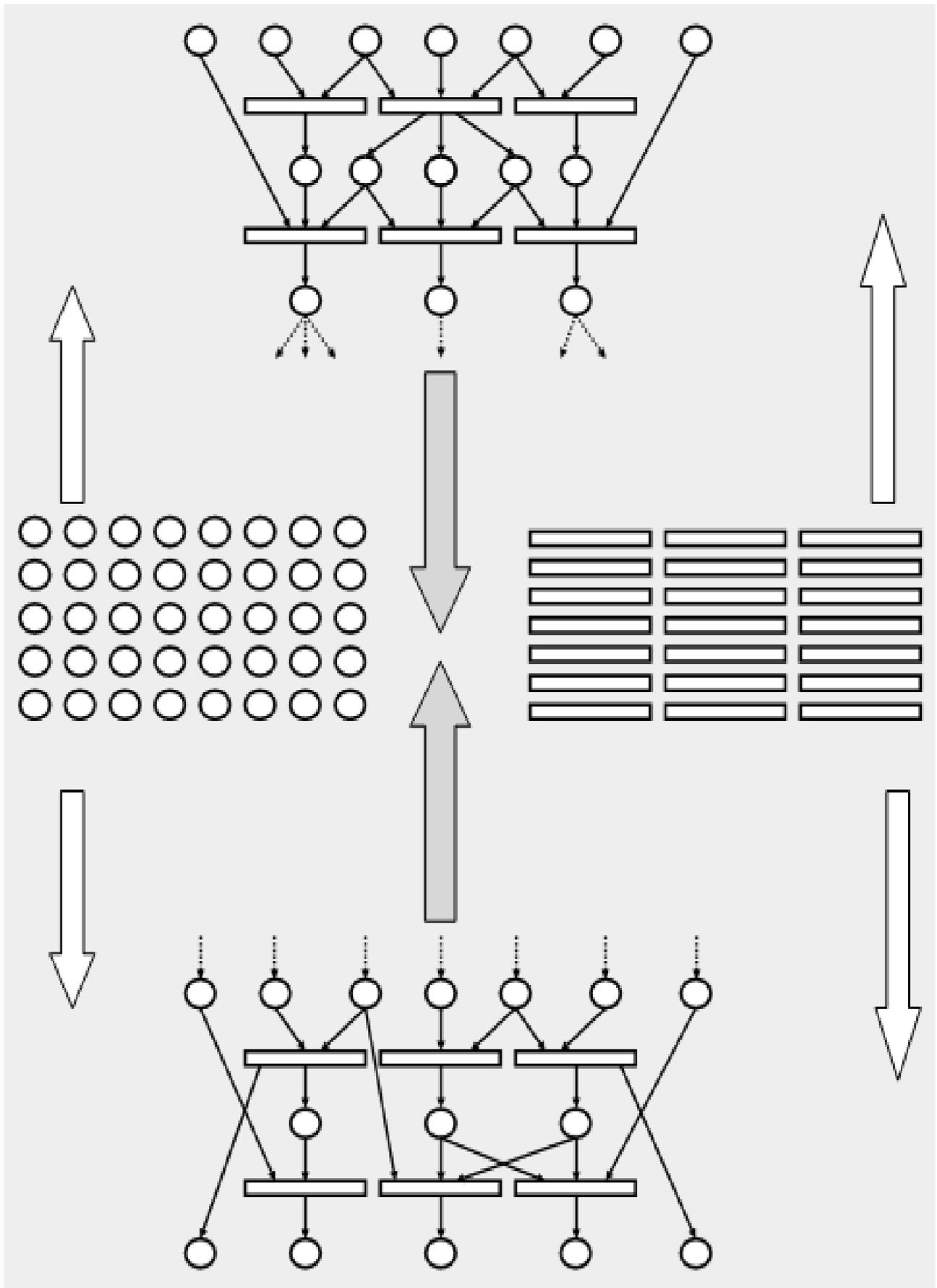

Picture 2. Representation of the MIVAR net in the form of bipartite graph.

```xml
<?xml version="1.0" encoding="UTF-8" ?>
-<root>
-<parametrs>
<parametr id="P1" value="0.0" description="... " />
<parametr id="P2" value="0.0" description="... " />
<parametr id="P3" value="0.0" description="... " />
...
</parametrs>
-<rules>
<rule id="R1" resultId="P1" initId="P2,P3" value="180-P2-P3" description="..." />
<rule id="R2" resultId="P2" initId="P1,P3" value="180-P1-P3" description="..." />
<rule id="R3" resultId="P3" initId="P1,P2" value="180-P1-P2" description="..." />
...
</rules>
-<metadata>
<idParametr inc="33" />
<idRule inc="161" />
</metadata>
</root>
```

Picture 3. An example of objects and rules in XML.

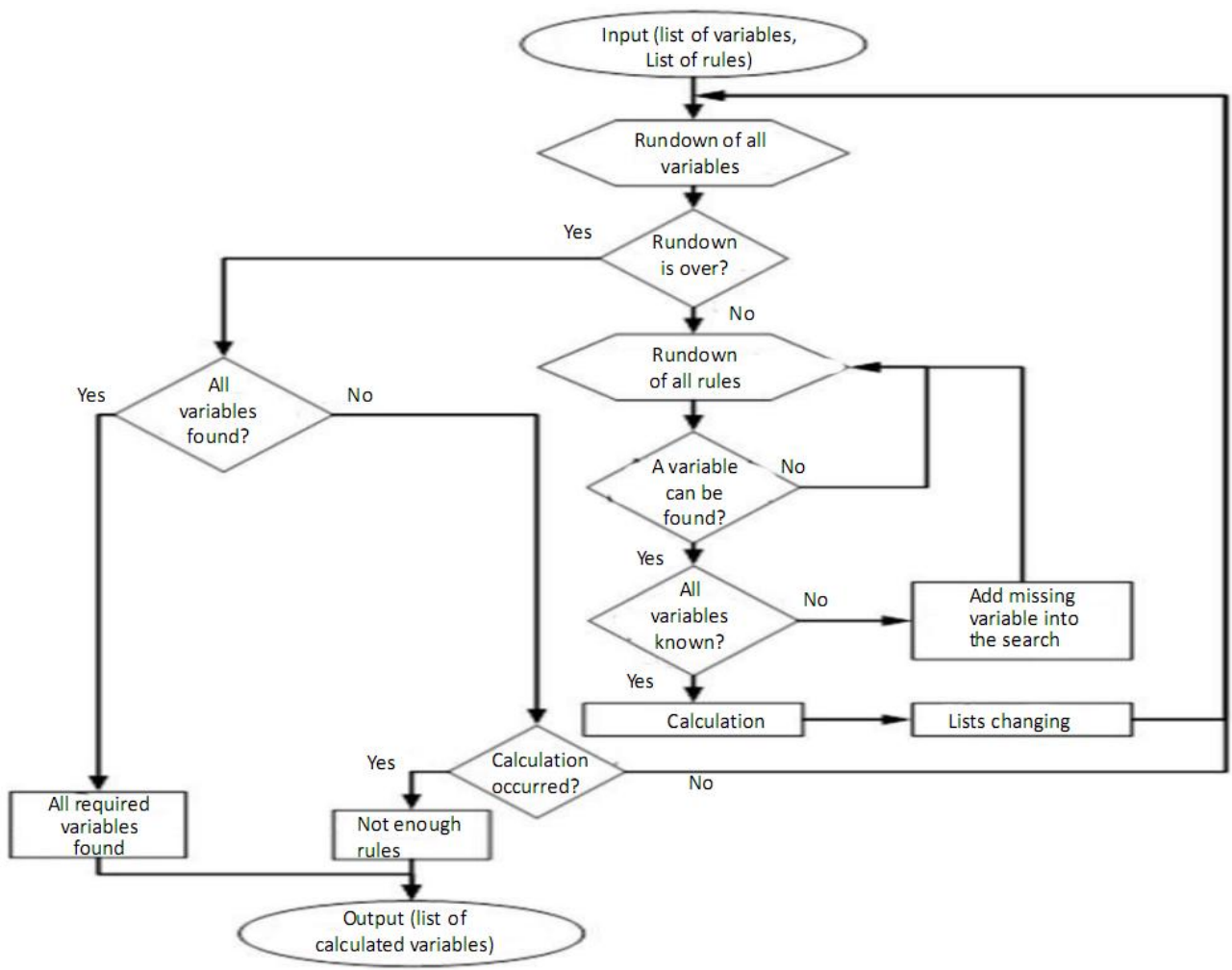

Picture 4. The algorithm of the UDAV work.

There are three basic stages of MIVAR data processing:

1) Creation of the MIVAR matrix for the description of subject domain;
2) Working with the matrix and the construction (designer) of the algorithm for solving of the required problem;
3) Execution of all computations basing on the acquired algorithm.

The first stage can be seen as the formalization of subject domain in the form of productions with the following transition to the MIVAR rules:

*"input objects – rules/procedures/services – output objects "*.

Currently, it is the most difficult stage that requires the participation of a human-specialist (expert) for creation of the MIVAR model of subject domain.

On the second stage, automated construction of the algorithm and logic inference is being implemented. The input data is represented in the form of the MIVAR matrix of the description of subject domain and specified input ("GIVEN") and required ("TO FIND") objects-variables.

On the third stage, the process of solution basing on the obtained algorithm is being executed. For the moment, in the software complex UDAV the work of the second and the third stages is combined. Currently, there are more than 7 different realizations of the MIVAR method. In some of them, three main stages are processed separately, but in this article the work of the UDAV software complex is described where all three stages can be combined.

Let's describe theoretic basis of the work of the MIVAR method of logic and computational data processing [2-4, 10, 15, 20]. It is to note that practical realizations in the form of concrete algorithms and software complexes may differ from the following general description.

First, a matrix for the MIVAR net of logic rules represented as a list is being constructed. Then, basing on the analysis of this matrix the fact of the existence of successful path of logic inference is being determined. On the last stage, the shortest of such paths is being chosen (the most optimal on the basis of the given criteria of optimality).

Let's consider that **m** rules and **n** variables are known (contained in the rules or input, activating them, or output). Then the matrix **V (m × n)** (each row of which corresponds to a rule and contains information about the variables used in this rule) contains all the relations about the rules and the variables. In each row, all the input variables of the rule on the corresponding positions are marked with **x**, all output – with **y**, all the variables that have already obtained some concrete value in the process of logic inference – with **z**, all required (output) – with **w**. In addition, let's use one more row and one more column to keep service information. Now we have the matrix **V (m+1) × (n+1)** that contains the structure of the net of rules. This structure can be change at any moment of time, so it is changeable, or evolutional. An example of such matrix is shown on the Pic.5. On the Pic.6 this matrix is shown in the form of the bipartite graph of the MIVAR net.

| V | 1 | 2 | 3 | 4 | 5 | ... | n-2 | n-1 | n | n+1 |
|---|---|---|---|---|---|-----|-----|-----|---|-----|
| 1 | X | X | X |   |   |     |     | Y | Y |   |
| 2 |   |   | X | Y | Y |     |     | X | X |   |
| ... |   |   |   |   |   | ... |     |     |   |   |
| m |   | X |   | X | X |     | Y |     |   |   |
| m+1 |   |   |   |   |   |     |     |     |   |   |

Picture 5. The initial matrix **V** of **(m+1) × (n+1)** with the structure of the net of rules.

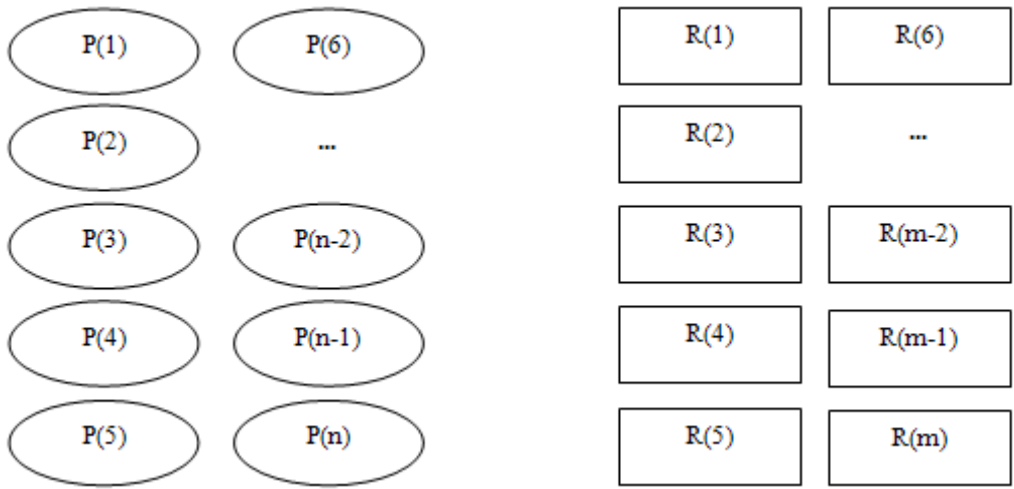

Picture 6. The initial bipartite graph of the MIVAR net (objects in the left column and rules in the right).

Let's describe an example of the work of the method. The following actions are being implemented to search the path of the logic inference.

1. In the row (m+1) all known variables are being marked with z, all required – with w. For example, on the Pic.7 positions 1, 2, 3 are marked with z, position (n-2) – with w. The Pic.8 shows the same action on the example of construction of the bipartite graph of the MIVAR net.

2. The search of the rules that can be activated (i.e. of those with all input variables known) is being implemented upside down. If there are no such rules the path of logic inference doesn't exist and the request for additional data should be sent. If there are such rules that can be activated, a corresponding mark in the service row of each of them is being done. For example, it is possible to mark them with the figure 1 (Pic.9, cell (1, n+1)). Pic.10 shows this action on the bipartite graph.

3. If there are several rules that can be activated, the choice of the rule to activate first is being implemented following previously determined criteria. Several rules can be processed at the same time provided that necessary resources are available.

4. The imitation of the launch of the rule (procedure) is being implemented by the assignment of the value "known" to the output variables of this rule (z in our example). Launched rule is being additionally marked with the figure 2 (not necessary). See an example on the Pic.11 and Pic.12.

5. If in the service row rest any required variables (w) the search of the path of logic inference is being continued. If there are no required variables, the problem is considered successfully solved, all launched rules composing the path of logic inference in the order they were launched. Pic.11 and 12 show only one launched rule, but there are still required variables, so it is necessary to proceed to the next step.

6. First, presence of those rules that can be launched after finding new values on the previous step is being determined. If there are no such rules the path of logic inference doesn't exist and the following actions are the same as on the step 2. If there are such rules the search of the path is being continued. In our example, there are such rules (see Pic.13). The cell (2, n+1) is marked with 1 to underline the possibility of the lunch of corresponding rule. Pic.14 shows the same on the bipartite graph.

7. The next step is similar to step 4. Then, following steps 5 and 6 all the actions are being repeated until the result is obtained. If it is necessary all the steps from 2 to 7 are being repeated until the result is obtained. The result may be positive – the path of logic inference exists, – and negative – there is no such path.  Let's continue our executing example step by step.

8. In the cells (m+1, 4) and (m+1, 5), the sign that variables 4 and 5 are deductible is obtained. The cell (2, n+1) is marked with 2, as the rule was launched. After that, we see that not all required variables are known. So, the processing of the matrix V (m+1) × (n+1) is to be continued. Analysis shows that the rule m can be launched (see Pic.17 and 18).

9. After the launch of the rule m new values are obtained (see Pic.19 and 20).

10. There are no more required rules in the service row. There are new values in the cells of the matrix: 2 in the cell (m, n+1) and z instead of w in the cell (m+1, n-2). So, the positive result is obtained, the path of logic inference exists.

| V | 1 | 2 | 3 | 4 | 5 | ... | n-2 | n-1 | n | n+1 |
|---|---|---|---|---|---|-----|-----|-----|---|-----|
| 1 | X | X | X | | | | | Y | Y | |
| 2 | | | X | Y | Y | | | X | X | |
| ... | | | | | | ... | | | | |
| m | | X | | X | X | | Y | | | |
| m+1 | Z | Z | Z | | | | W | | | |

Picture 7. Step 1 on the matrix **V.**

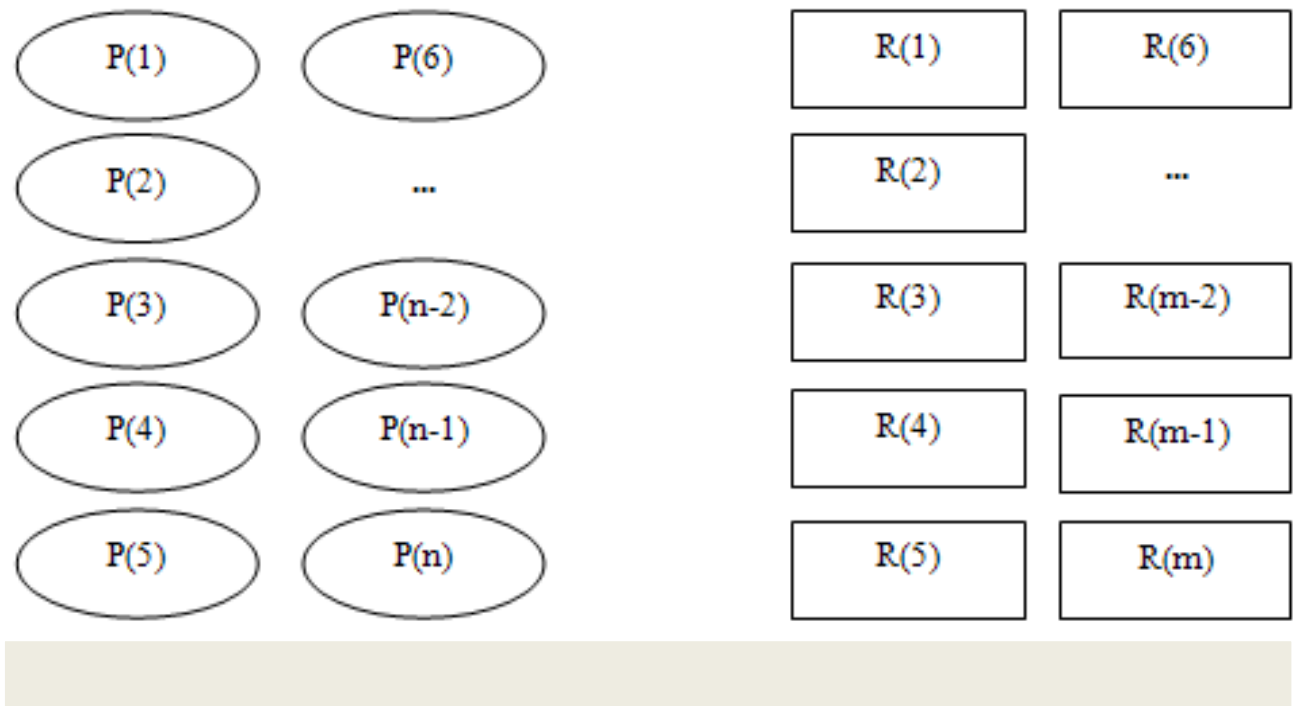

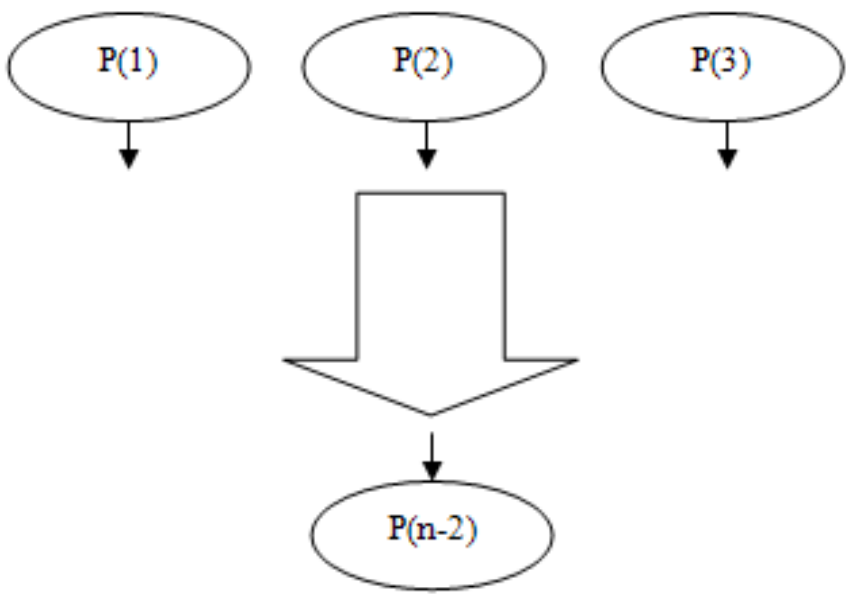

Picture 8. Step 1 on the bipartite graph.

| V | 1 | 2 | 3 | 4 | 5 | ... | n-2 | n-1 | n | n+1 |
|---|---|---|---|---|---|-----|-----|-----|---|-----|
| 1 | X | X | X | | | | | Y | Y | **1** |
| 2 | | | X | Y | Y | | | X | X | |
| ... | | | | | | **...** | | | | |
| m | | X | | X | X | | Y | | | |
| m+1 | Z | Z | Z | | | | W | | | |

Picture 9. Step 2 on the matrix **V.**

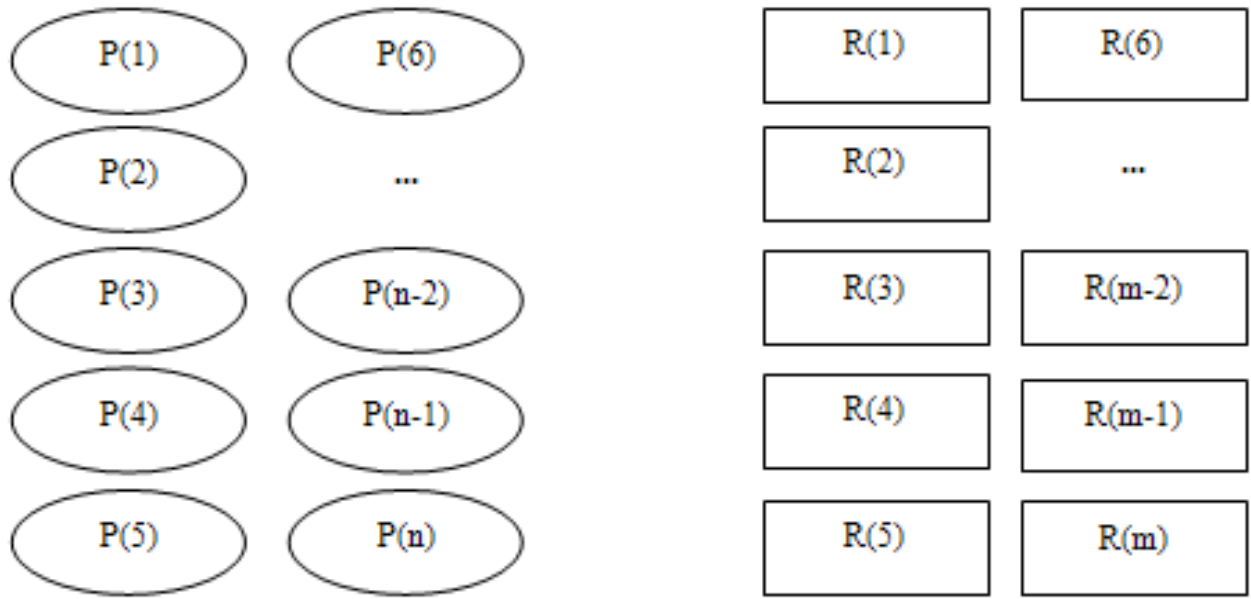

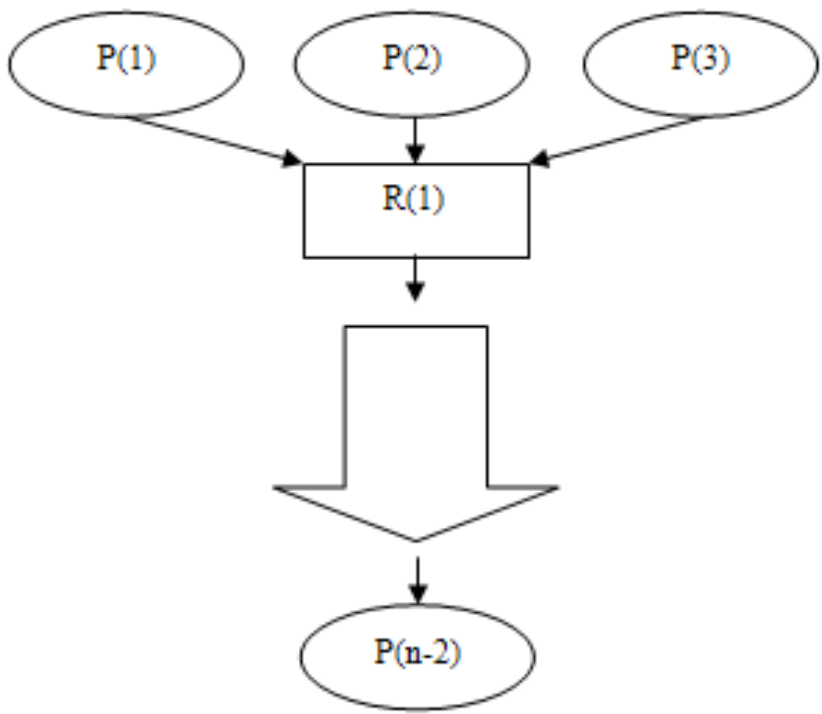

Picture 10. Step 2 on the bipartite graph.

| V | 1 | 2 | 3 | 4 | 5 | ... | n-2 | n-1 | n | n+1 |
|---|---|---|---|---|---|-----|-----|-----|---|-----|
| 1 | X | X | X | | | | | Y | Y | 2 |
| 2 | | | X | Y | Y | | | X | X | |
| ... | | | | | | ... | | | | |
| m | | X | | X | X | | Y | | | |
| m+1 | Z | Z | Z | | | | W | Z | Z | |

Picture 11. Step 3 on the matrix **V.**

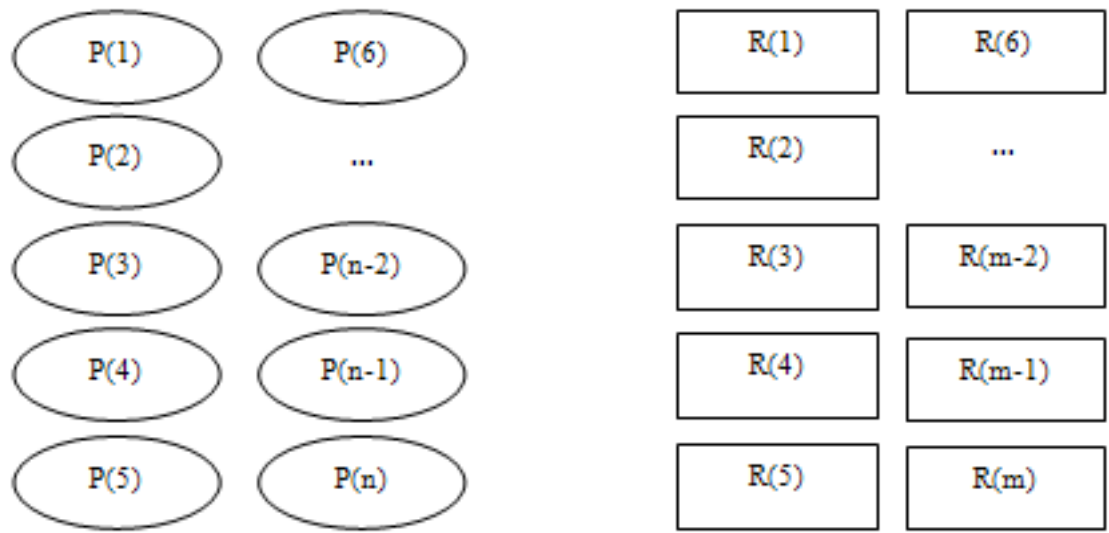

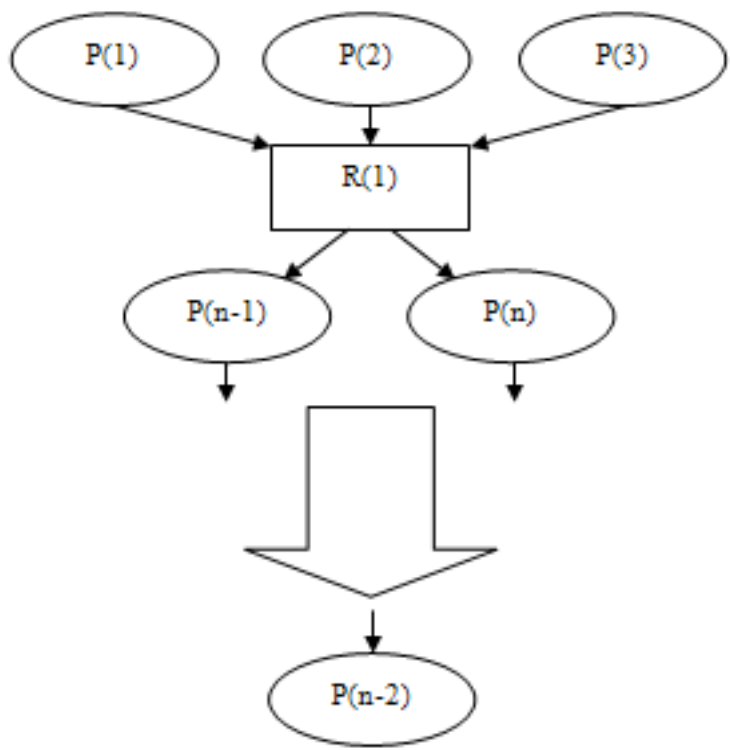

Picture 12. Step 3 on the bipartite graph.

| V | 1 | 2 | 3 | 4 | 5 | ... | n-2 | n-1 | n | n+1 |
|---|---|---|---|---|---|-----|-----|-----|---|-----|
| 1 | X | X | X |   |   |   |   | Y | Y | 2 |
| 2 |   |   | X | Y | Y |   |   | X | X | 1 |
| ... |   |   |   |   |   | ... |   |   |   |   |
| m |   | X |   | X | X |   | Y |   |   |   |
| m+1 | Z | Z | Z |   |   |   | W | Z | Z |   |

Picture 13. Step 4 on the matrix **V.**

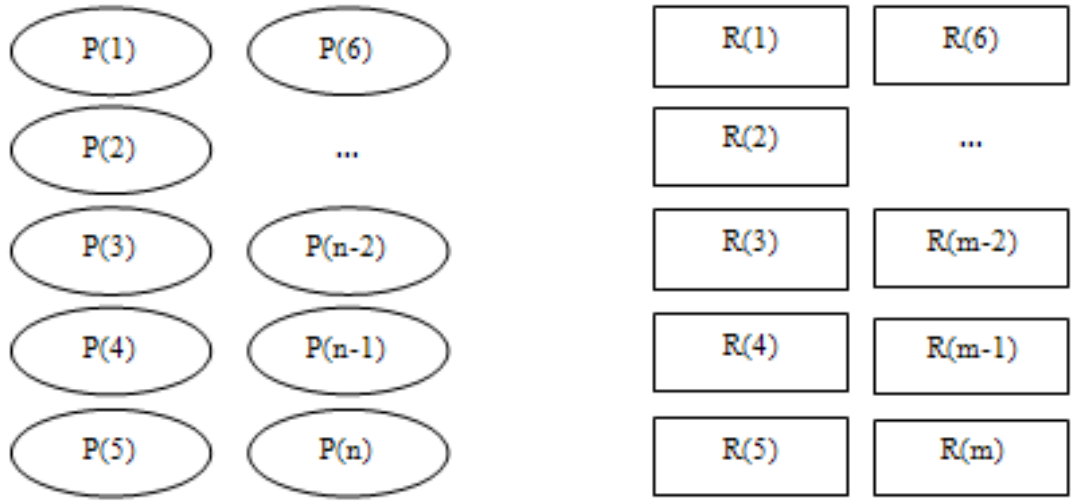

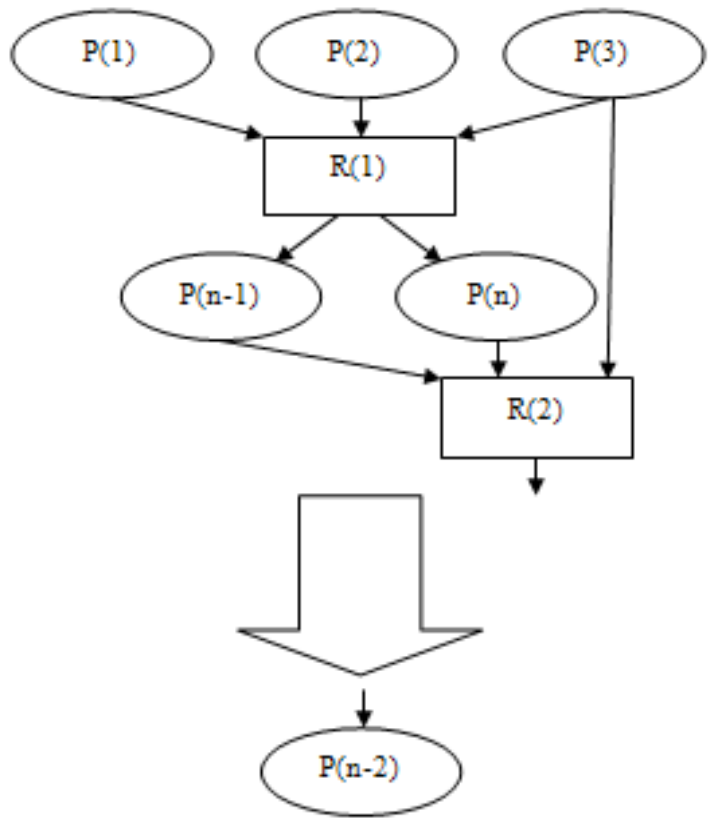

Picture 14. Step 4 on the bipartite graph.

| V | 1 | 2 | 3 | 4 | 5 | ... | n-2 | n-1 | n | n+1 |
|---|---|---|---|---|---|---|---|---|---|---|
| 1 | X | X | X | | | | | Y | Y | 2 |
| 2 | | | X | Y | Y | | | X | X | 2 |
| ... | | | | | | ... | | | | |
| m | | X | | X | X | | Y | | | |
| m+1 | Z | Z | Z | Z | Z | | W | Z | Z | |

Picture 15. Step 5 on the matrix **V.**

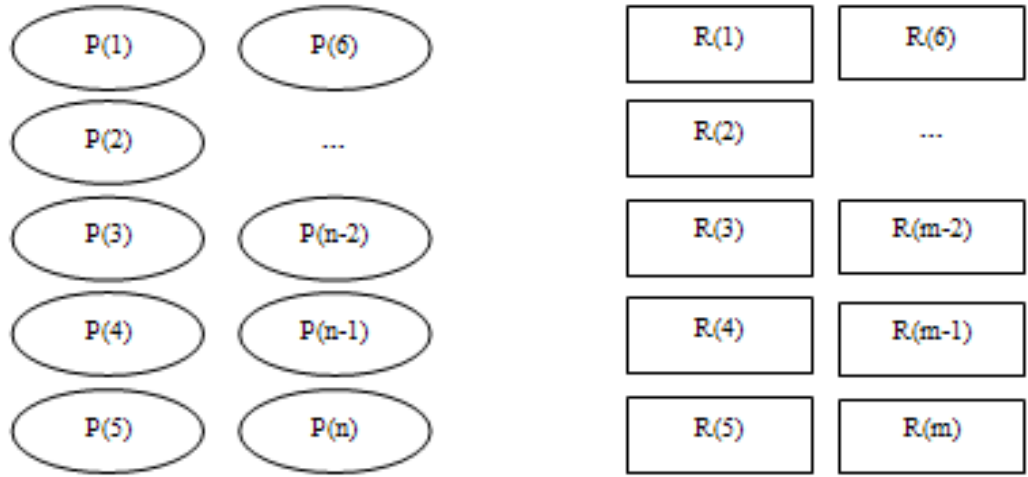

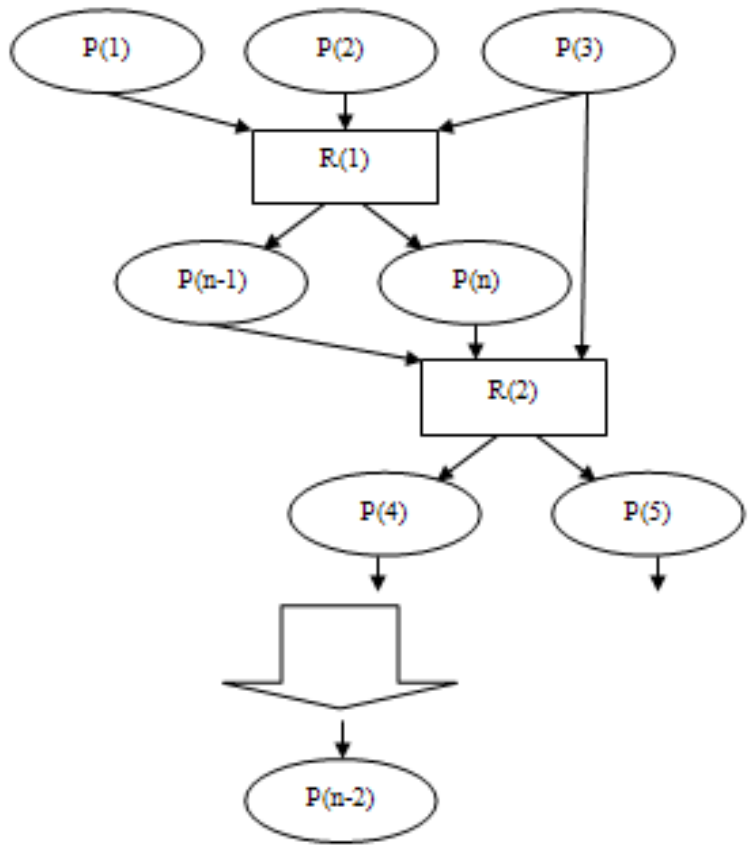

Picture 16. Step 5 on the bipartite graph.

| V | 1 | 2 | 3 | 4 | 5 | ... | n-2 | n-1 | n | n+1 |
|---|---|---|---|---|---|-----|-----|-----|---|-----|
| 1 | X | X | X | | | | | Y | Y | 2 |
| 2 | | | X | Y | Y | | | X | X | 2 |
| ... | | | | | | ... | | | | |
| m | | X | | X | X | | Y | | | 1 |
| m+1 | Z | Z | Z | Z | Z | | W | Z | Z | |

Picture 17. Step 6 on the matrix **V.**

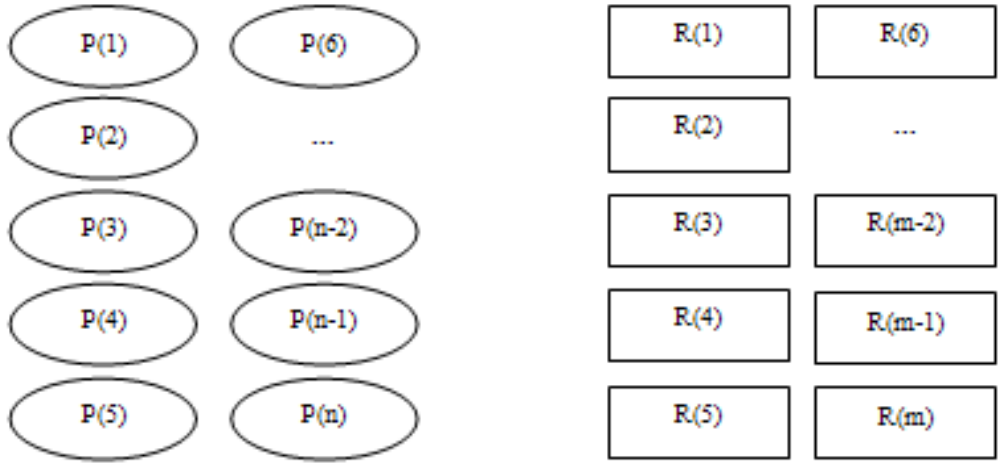

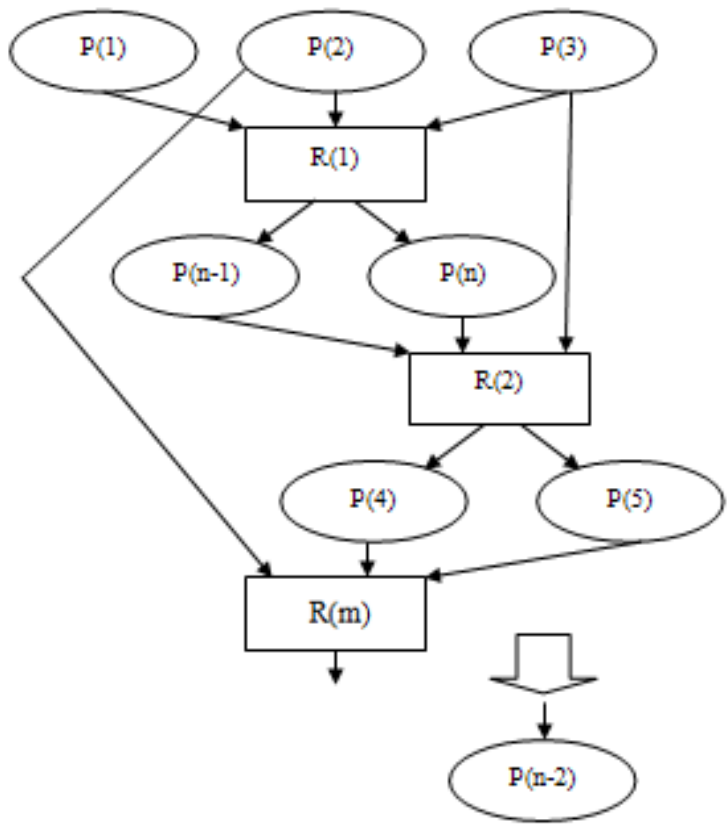

Picture 18. Step 6 on the bipartite graph.

| V | 1 | 2 | 3 | 4 | 5 | ... | n-2 | n-1 | n | n+1 |
|---|---|---|---|---|---|---|---|---|---|---|
| 1 | X | X | X | | | | | Y | Y | 2 |
| 2 | | | X | Y | Y | | | X | X | 2 |
| ... | | | | | | ... | | | | |
| m | | X | | X | X | | Y | | | 2 |
| m+1 | Z | Z | Z | Z | Z | | Z(W) | Z | Z | |

Picture 19. Step 7 on the matrix **V.**

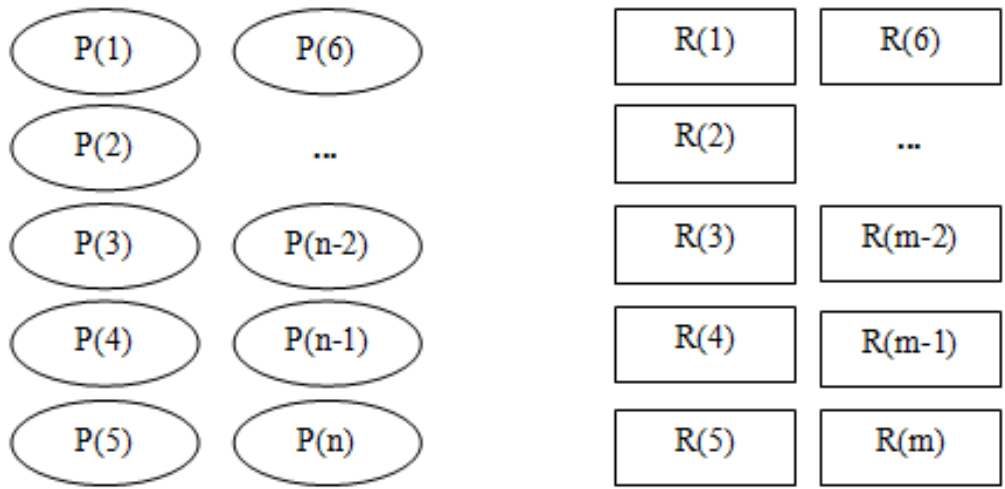

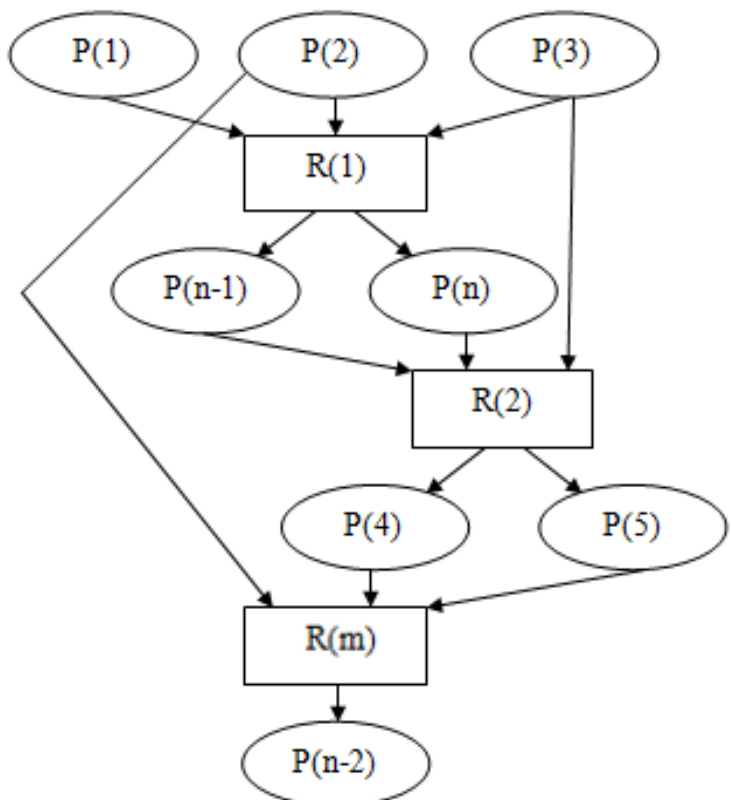

Picture 20. Step 7 on the bipartite graph.

**Advantages of the MIVAR method**

It has been shown that bipartite graphs of MIVAR nets can be used for expert systems on the basis of production. From scientific point of view, the main difficulty in using expert systems is conceptual description of subject domain in terms of productions and the formation of two necessary lists: objects and rules for the MIVAR logic nets. The processing itself is a universal mechanism described in [1-22]. Rules-procedures can be represented as productions corresponding to traditional approach. Universal possibilities of the MIVAR approach are caused by the fact that it unifies all known data models, including the model "entity-relation", Petri nets, semantic models and ontology [4, 10, 15, 18-20]. The advantages of the MIVAR approach are:

1) lineal computational complexity and real time working;
2) solution of logic and computational (and other) tasks;
3) input data flow control and data-driven search;
4) adaptive description and ongoing problem solving;
5) active work with the requests or redeterminations of input data.

**Theoretical estimation of the MIVAR method computational complexity**

The total number of operations used in the MIVAR method is determined by the sum of the actions at each stage:

1) assignment of known z and required w to the cells of the service row (m+1) (total number of operations no more than n);

2) assignment of the sign of the processing of rules in the service column (n+1) (total number no more than 2m but can be no more than m);

3) assignment of the sign that the variable is known (z) to the cells of the service row (m+1) (total number no more than n);

4) assignment of the new values to the cells of the row (m+1) (no more than n operations).

Operations on the steps 1,3 and 4 are accomplished over the row (m+1). The total number of actions isn't more than the total number of the cells in this row, as already processed values are "erased" and not processed again. In the whole, the total number of operations (**KD**) in the MIVAR method, i.e. its computational complexity doesn't overwhelm the number of cells in service parts of the matrix:

$$O(m+n), \text{т.е. } KD \leq (m+n)$$

In the worst (complicated) case when the proposed reduction of the computations can't be realized computational complexity of this method is:

$$KD = O(mn),$$

Papers [12-15, 20-22] and web resources [18-19] show the results of the practical realization of the MIVAR nets in the UDAV software complex. Thus, papers [1-22] prove that the MIVAR approach is universal for the solution of various practical problems with simultaneous logical and computational data processing. So, the contradiction between logic inference and computational processing is successfully overcome by the means of the MIVAR nets.

**Results of experiments**

Currently, there are several software complexes – prototypes of the expert systems that implement the MIVAR method of data processing on practice. The most important of them is UDAV software complex [9-22] which is the base for the project of "Active MIVAR encyclopaedia". The MIVAR encyclopaedia describes three subject domains: "triangles", "stereometry" and "circumferences". G.S. Sergushin has independently implemented a software complex "physics of falling body" [10, 18-19] and a prototype for the MIVAR expert system of the management of car engine. Also, a prototype of expert system – simulator of electric substation has been implemented. This simulator allows its user to reconfigure the structure of the logic scheme of electric substation and to add new elements.

As mentioned above, the most difficult part in the MIVAR method is the description of subject domain in the form of MIVAR net. To accomplish numerical experiments, UDAV software complex was created. New test version of this complex includes a generator of the MIVAR matrixes for the description of subject domains. This feature was added on order to create matrixes of big size. For example, matrixes containing 1 170 007 objects and 3 510 015 rules, i.e. 3,5 million of cells were created during an experiment. To create such matrixes, a specially elaborated algorithm is used.

The input of this algorithm is the number of objects-variables. Then, the algorithm generates simple arithmetic rules for a given number of objects. For example, such rules may be the following:

$$a+b=c \quad \text{and/or} \quad c-b=a \quad \text{and/or} \quad c-a=b$$

The algorithm accomplishes matrix generation and saves it in a special file. Then, some test programme reads test matrix and begins to process it. For all executed comparisons to be single-valued and adequate all generated matrixes are similar. Their similarity is provided to be the algorithm of simple rules generation and the assignment of control rules. Objects $P1=10$ and $P2=10$ are always being chosen as known objects. Object under the last number is always being chosen as required object. Thus, if the number of input parameters is 100 the algorithm will search the object number $P100$, if the number of input parameters is 10000 it will search $P10000$.

The user can generate his own matrixes or change already created. Such experiments have been held. The test programme is written only to evaluate the working time of UDAV kernel with elementary rules and proportional increase of MIVAR matrixes.

In the end of its work, the test programme builds the solution graph where the parameters are marked with pink colour (with the identifier inside) and the rule is marked with orange colour (also with the identifier inside). The input data is on the upper part of the graph, and the required parameter is in the bottom part. Generator is designed in the way that the solution always exists, and to find this solution proportional increase of the number of algorithm steps is necessary.

Working time of the programme may depend on general CPU and simultaneously executed tasks. Currently, both single-flow and multi-flow modes of work are realized. The major part of experiments has been accomplished in the single-flow mode. The test programme considers neither the time of test matrix creation nor the time of creation of the solution graph in the end.

The time of solution is given in milliseconds (ms). Experiments showed that RAM resource is the main limitation of the number of processed objects.

Tests were held on different computers beginning from netbooks with RAM 512 Mb to small servers with Intel processor of 3,8 GHz and RAM 4 GB. The number of kernels did not affect working speed, as all main processes were executed using the single-flow programme. Windows and MacOs were used.

When Windows operating systems were used the following results were obtained: up to 150000 objects and 450000 rules on the computers with RAM 4 HB. The time of solution was from several ms to 2 793 672 ms which is about 47 minutes depending on the CPU.

The results of experiments carried out on the MacBook with the OS MacOSX 10.6.7 and Intel Core 2DUO processor working at 2GHz and with RAM DDR3 of 4HB are shown on the Pic.21. Figure shows the time on a vertical solution, in milliseconds, and the number of rules, and the horizontal shows the number of objects MIVAR net. On the basis of the MIVAR nets the UDAV software complex is developed, handling more than 1.17 million objects and more than 3.5 million rules on ordinary computers.

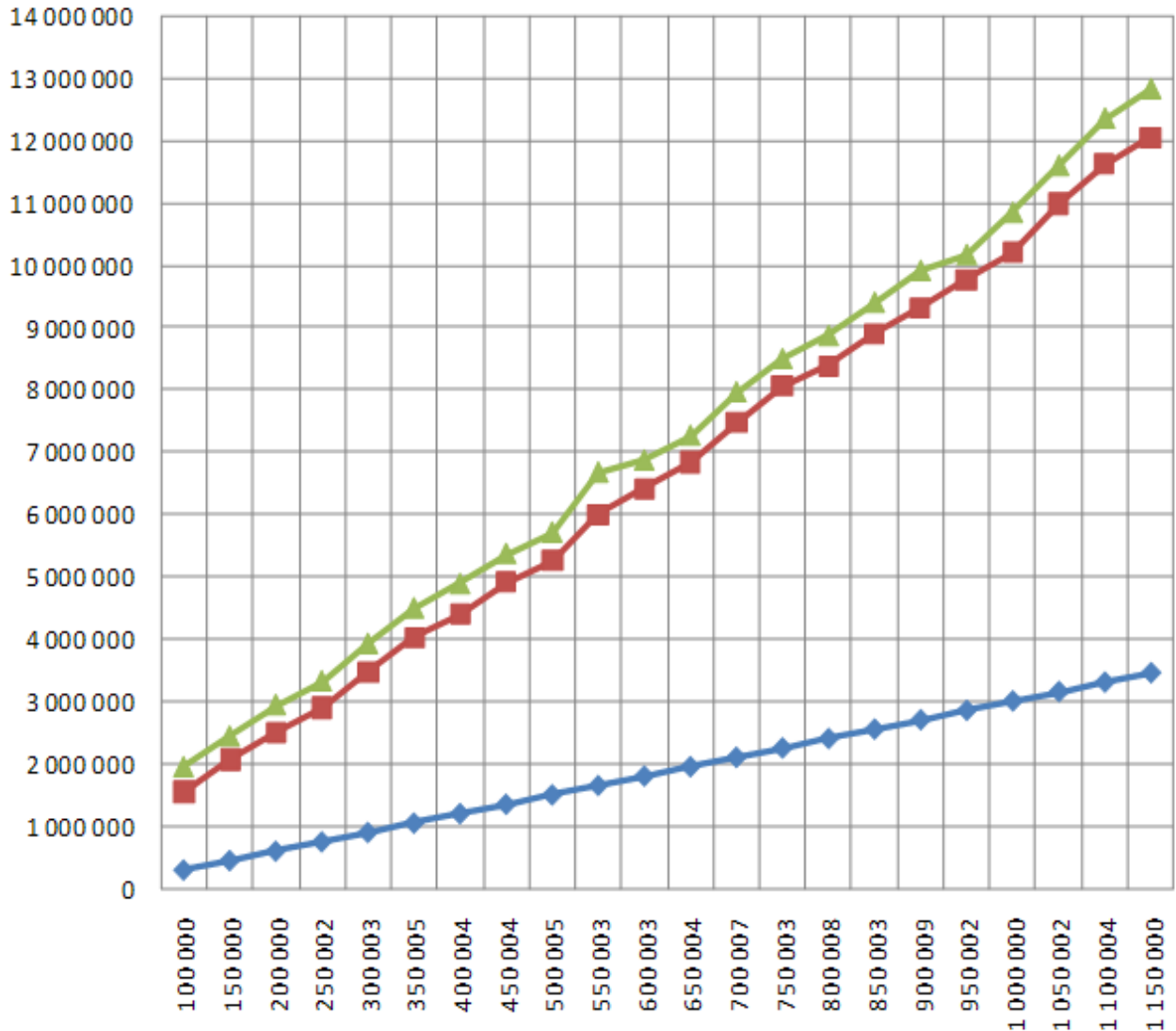

Picture 21. Results of experiments (time of solution from the number of objects and rules: figure shows the time on a vertical solution, in milliseconds, and the number of rules, and the horizontal shows the number of objects MIVAR net).

The test program written on JAVA process MIVAR matrixes of 1 million objects and 3.5 million parameters in 3 hours. The maximal acquired figures are the following: 1 170 000 objects-variables, 3 510 015 rules, the time of solution 12 239 183 ms or about 200 minutes. To compare with, the system of control under big submarines process about 20 000 rules; the description of all school programme and technical institute can be described in about 300 rules for each scientific domain and no more than 100 000 in general. At last, traditional production and predicate systems couldn't work with a hundred of rules.

Thus, the possibilities of the UDAV software complex and the MIVAR method of information processing allow creating of global expert system that would unify all formalized knowledge of mankind. This is the project "MIVAR active encyclopaedia".

In general, the experiments were carried out on more than 15 computers. For each number of objects-variables more than 10 similar trials under the nearly same conditions were held. The calculation showed that the error in time of programme work measurement is no more than 3%. All experiments proved in practice the lineal computational complexity of the MIVAR method.

## Conclusion

1.  The theoretical transition from the graphs of production systems to the bipartite graphs of the MIVAR nets is shown. Examples of the implementation of the MIVAR nets in the formalisms of matrixes and graphs are given. The linear computational complexity of algorithms for automated building of objects and rules of the MIVAR nets is theoretically proved. Different services, modules and computational procedures can be used as the MIVAR rules. The MIVAR method of data processing is fully described. It is proved that the bipartite graphs of the MIVAR nets can be used in the expert systems instead of productions and traditional graphs.

2.  On the basis of the MIVAR nets the UDAV software complex is developed, handling more than 1.17 million objects and more than 3.5 million rules on ordinary computers. The results of experiments that confirm a linear computational complexity of the MIVAR method of information processing are given.

3.  UDAV can be used for the search of logic inference. MIVARs allow creating a "General Problem Solver" the prototype of which is UDAV. Active logically thinking systems basing on the MIVARs can be created which will be the prototype of Artificial Intelligence.

4.  The possibilities of creating of multi-subject expert systems. In the perspective, a global multi-subject active expert system "The MIVAR active encyclopaedia" will be created basing on the MIVAR nets.

5.  The MIVAR nets eliminate previous limitations. MIVAR technologies of the data storage and processing expand computers technical possibilities and allow creating new intellectual systems. The MIVAR nets are the qualitative leap to the new possibilities of data processing.